\documentclass[11pt]{article}

\usepackage[final]{coling}

\usepackage{times}
\usepackage{latexsym}
\usepackage{float}

\usepackage[T1]{fontenc}

\usepackage[utf8]{inputenc}

\usepackage{microtype}

\usepackage{inconsolata}

\usepackage{graphicx}

\title{Romanized to Native Malayalam Script Transliteration Using an Encoder-Decoder Framework}

\author{Bajiyo Baiju, Kavya Manohar, Leena G Pillai, Elizabeth Sherly \\
       Digital University Kerala \\ Thiruvananthapuram \\ Kerala, India}

\begin{document}
\maketitle
\begin{abstract}
In this work, we present the development of a reverse transliteration model to convert romanized Malayalam to native script using an encoder-decoder framework built with attention-based bidirectional Long Short Term Memory (Bi-LSTM) architecture. To train the model, we have used curated and combined collection of 4.3 million transliteration pairs derived from publicly available Indic language translitertion datasets, \textit{Dakshina} and \textit{Aksharantar}. We evaluated the model on two different test dataset provided by \textit{ IndoNLP-2025-Shared-Task} that contain, (1) General typing patterns and (2) Adhoc typing patterns, respectively. On the Test Set-1, we obtained a character error rate (CER) of 7.4\%. However upon Test Set-2, with adhoc typing patterns, where most vowel indicators are missing, our model gave a CER of 22.7\%.

\end{abstract}

\section{Introduction}



Typing in native script has always remained a challenge for speakers of many Indian languages including Malayalam, across diverse digital platforms. In the pre-smartphone era, where native language typing was virtually non-existent due to the unavailability of accessible and user-friendly keyboards, typing Malayalam in the Roman script was the norm. Even with advancements in technology, typing in the Roman script has become the natural and preferred mode of input across devices for an average user \cite{madhani-etal-2023-aksharantar}. While romanized communication seems convenient, it is not preferred in formal contexts.

Transliteration from romanized input to native scripts is inherently complex due to variations in typing styles, the absence of standardized romanization schemes, and the context-dependent nature of character mappings. Hence there is a need for real-time transliteration tools that can seamlessly convert romanized Malayalam into its native script. In this work, we address this need by developing a robust reverse transliteration model for Malayalam, where romanised Malayalam is automatically converted into native script.

The proposed model leverages an attention-based bidirectional Long Short Term Memory (Bi-LSTM) encoder-decoder framework, trained on large-scale transliteration datasets, namely \textit{Dakshina} \cite{roark-etal-2020-processing} and \textit{Aksharantar} \cite{madhani-etal-2023-aksharantar}. The code for training the model is published under MIT License \footnote{\url{https://github.com/VRCLC-DUK/ml-en-transliteration}}. This paper outlines the related works, datasets, model architecture and results, highlighting the model's performance on datasets that reflect both general and adhoc typing patterns.

\section{Related Works}

Rule-based and data-driven approaches are the two
main strategies for transliteration \cite{kavya2022}. Prior to the advent of deep learning approaches of learning from huge data, rule based approaches were the norm. In the context of well defined romanization standards \cite{iso}, rule based approaches are the best in terms of speed and accuracy. However there are non-standard romanised Malayalam used in informal communication contexts, that calls for deep learning solutions.

A rule based system available for transliteration among Indian languages based on soundex algorithms is introduced in \textit{Libindic} \cite{libindic}. \textit{Aksharamukha} script converter is is another rule-based systems that transliterates among 121 scripts and 21 standard romanization methods \cite{aksharamukha}. The \textit{Brahmi-Net} tool covers 306 language pairs across 13 Indo-Aryan, 4 Dravidian languages, and English, utilizing an unsupervised method to mine parallel transliteration corpora for statistical training. This hybrid system leverages Unicode ranges and an extended ITRANS encoding to enable script conversions between Brahmi-derived scripts \cite{kunchukuttan-etal-2015-brahmi}.

Deep learning approaches rely on carefully crafted transliteration corpora for training the models. \textit{Dakshina} is an open licensed and curated transliteration corpora \cite{roark-etal-2020-processing} consisting of native script text, a romanization lexicon and some romanized full sentences in 12 south Asian languages. \textit{Aksharantar} is the largest publicly available transliteration dataset with 26 million transliteration pairs for Indian languages created by mining from  monolingual and parallel corpora, as well as collecting data from human annotators \cite{kunchukuttan-etal-2021-large,madhani-etal-2023-aksharantar}. It has also been reported that mined name pair datasets \cite{namepair} could be used for training general purpose transliteration models \cite{Baiju2024}. 

A multitask learning based training for multilingual neural transliteration leveraging orthographic similarity between languages was described in \cite{kunchukuttan2018leveraging}. Non-neural method like pair \textit{n-gram} and nueral methods like sequence-sequence LSTM  and transformer architectures were compared in \citeauthor{roark-etal-2020-processing} for single words transliteration task.
Transliteration implemented using neural machine translation system (NMT) was proposed by \citeauthor{kunchukuttan-etal-2021-large}, where \textit{Marian} \cite{junczys-dowmunt-etal-2018-marian} was used for training the model. IndicXlit is a multilingual neural transliteration model trained on the  \textit{Aksharantar} \cite{madhani-etal-2023-aksharantar} dataset using an encoder-decoder transformer architecture. Grapheme to Phoneme Conversion systems for mapping of Malayalam script to precise romanisation schemes have been explored in rule based \cite{baby2016resources,parlikar2016festvox, manghat2020malayalam} and data driven \cite{Priyamvada_2021} fashions.

Transliterating sentences are considered as a different task than transliterating single words in \cite{roark-etal-2020-processing}. Identifying word contexts can improve sentence level transliteration. \citeauthor{kirov2024context} describes methods to incorporate language models to improve transliteration of full sentences as opposed to single words.

\section{Methodology}

The methodology involved in this study encompasses the curation and preprocessing of training datasets, design of the model architecture, training, and evaluation on the test data set. 

In the current work, we train word-level transliteration model. During testing, we preprocess sentences by extracting individual words, performing word-level transliteration, and then reconstructing the full sentence in the post-processing stage. Non-alphabetic characters like punctuation and numbers are excluded from model input, preserved in their original positions, and reinserted after generating the transliterated token sequence.

\subsection{Datsets}

\begin{table*}[ht]
  \caption{An illustration of 3 character errors distributed across 3 words severly deteriorating WER and BLEU scores. The errors in transliteration are indicated in red color.}   \label{tab:test1}
  \centering
   \begin{tabular}{c}
       \includegraphics[scale=0.8]{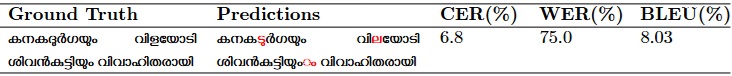} \\ 
  \end{tabular}
\end{table*}




The reverse transliteration model for romanized Malayalam is trained on two publicly available curated collection of Indic language transliteration datasets: \textit{Dakshina}\footnote{\url{https://github.com/google-research-datasets/dakshina}} and the \textit{Aksharantar}\footnote{\url{https://huggingface.co/datasets/ai4bharat/Aksharantar}}. The \textit{Dakshina} dataset comprises of 244 thousand single word transliteration pairs, while the \textit{Aksharantar} dataset adds a significantly larger volume of 4.100 million pairs. Together, these datasets comprise a total of 4.344 million word-level transliteration pairs. The combined dataset ensures a rich and diverse training set that includes both common and rare transliteration patterns, capturing variations in typing styles and phonetic representations.


Each entry in the dataset is structured as a pair of columns: `ml' and `en'. The `ml' column represents the native Malayalam script, while the `en' column contains the corresponding romanized representation. This consistent and simple structure facilitates efficient preprocessing and model training, enabling the encoder-decoder framework to learn the mapping between the romanized input and the native script output effectively.

\subsection{Model Architecture and Training}
The proposed reverse transliteration model for converting romanized Malayalam to native script is based on an attention-enabled encoder-decoder framework utilizing Bi-LSTM layers. We define separate source and target tokenizers. The source tokens are lower case Latin characters and the target tokens are Malayalam characters comprising of vowels, vowel signs, consonants and the special characters \textit{anuswaram}, \textit{visargam}, \textit{virama} and \textit{chillu} \cite{kavya2022}.

\begin{figure*}[!htpb]
  \includegraphics[width=\textwidth]{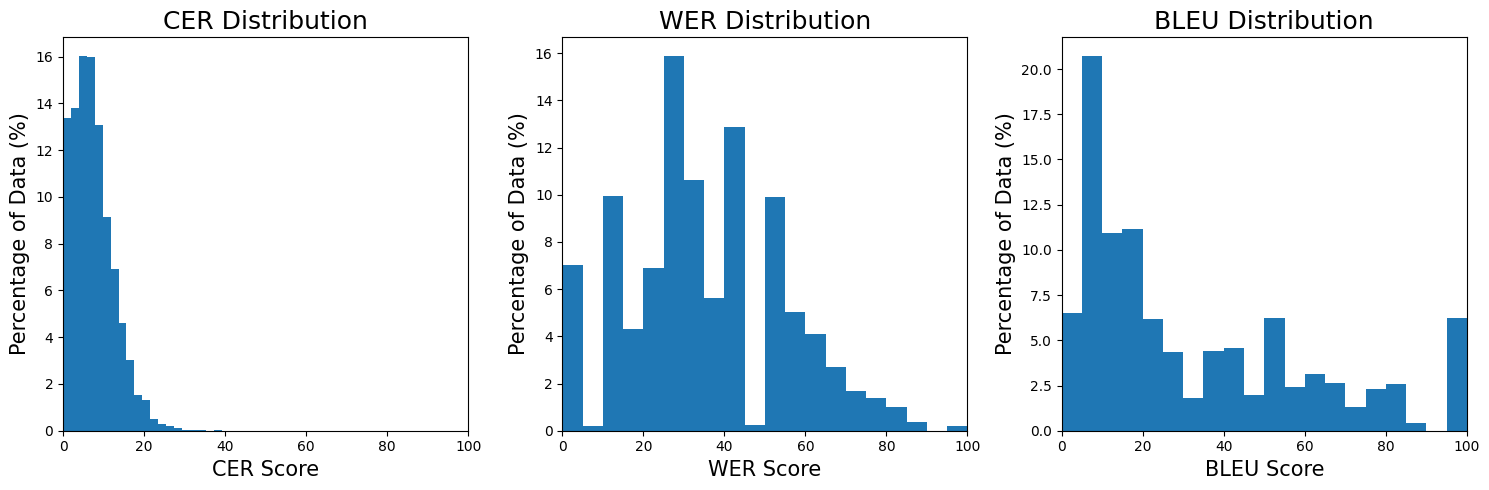}
  \caption{The distribution of WER, CER and BLEU over the Test Set-1. }
  \label{fig:test1}
\end{figure*}

\begin{figure*}[!htpb]
  \includegraphics[width=\textwidth]{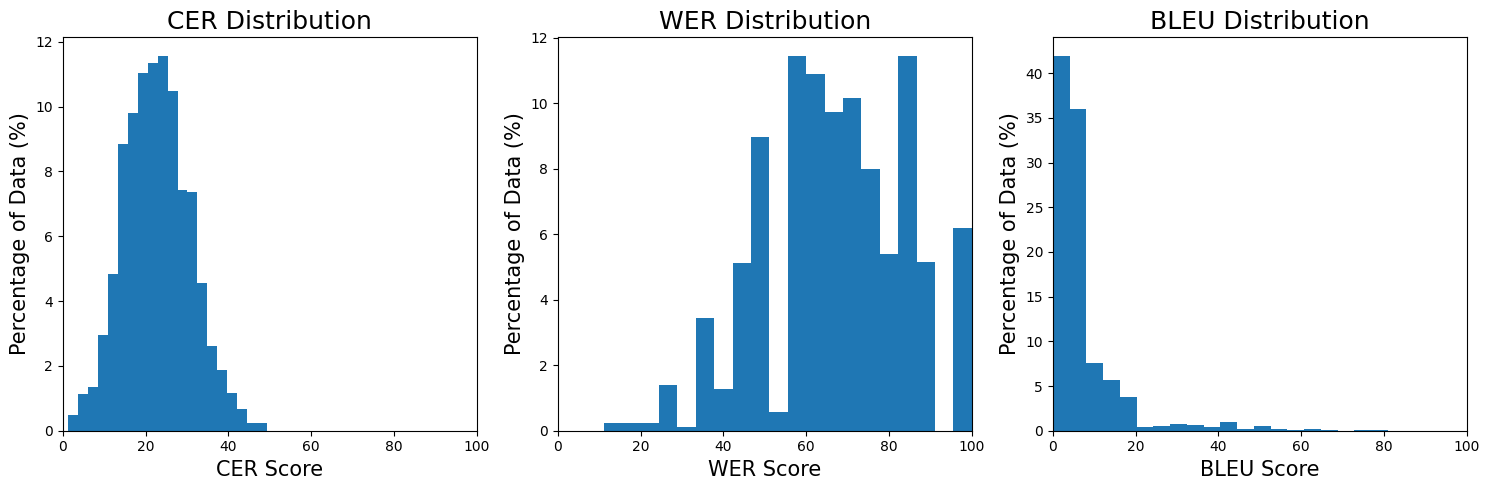}
  \caption{The distribution of WER, CER and BLEU over the Test Set-2.}
  \label{fig:test2}
\end{figure*}

The architecture begins with the encoder input layer, which accepts input sequences of up to 57 characters, which is identified as a maximum input sequence length from the training data. These sequences are integer-encoded representations of characters, serving as the foundation for subsequent layers. The next step involves the embedding layer, which transforms each character in the sequence into a 64-dimensional dense vector which allows the model to capture semantic relationships among characters in a continuous space.

Following this, a bidirectional LSTM layer processes the embedded input sequences to capture information from both past and future characters in the sequence. The bidirectional output, consisting of hidden states from both directions, is concatenated to form a 256-dimensional representation for each timestep. To reduce dimensionality and adjust the feature representation, a dense layer is applied, resulting in a 128-dimensional vector for each timestep. The context vector extracts from this processed sequence, and it serves as the initial input to the decoder.

The decoder begins with the repeat vector layer, which duplicates the context vector for each timestep of the target sequence. This ensures that all decoder timesteps have access to the same initial context. The repeated context vector is processed by an LSTM layer in the decoder, which generates a sequence of hidden states by modeling temporal dependencies in the target sequence. These states form the basis for the generation of the transliterated output. The model incorporates an attention mechanism (attention layer) to enhance its ability to focus on relevant parts of the input sequence during decoding. 

The output of the LSTM decoder and the attention layer is concatenated to form a unified representation, combining temporal dependencies with context-aware features. This enriched representation is passed through a time-distributed dense layer, which applies a dense transformation to each timestep. The result is a sequence of probability distributions over the 76 output characters, from which the final transliterated word is constructed.  A single Nvidia DGX A100 GPU with 80 GB RAM was used for training the model.

\section{Results}

\begin{table}[htpb]
  \caption{Evaluation metrics averaged over respective test datsets}   \label{tab:final}
  \centering
   \begin{tabular}{lccc}

    \hline
    \textbf{Dataset} & \textbf{CER} & \textbf{WER} &\textbf{BLEU}\\
    & \textbf{(\%)} & \textbf{(\%)} &\textbf{(\%)}\\
    \hline
Test Set-1 & 7.4 &  34.5 & 32.7 \\ \hline
Test Set-2 & 22.7 & 66.9 & 7.5 \\ \hline

    \hline
  \end{tabular}

\end{table}

We evaluate our model's performance using the IndoNLP Shared Task dataset\footnote{\url{https://github.com/IndoNLP-Workshop/IndoNLP-2025-Shared-Task}} for Malayalam. The test set is divided into two categories: Test Set-1, which includes general transliteration patterns, and Test Set-2, which features adhoc transliteration patterns where the romanized text omits several vowels.  These datasets consist of sentence-level samples. Samples of ground truth and predicted samples in test sets are linked in the repository\footnote{\url{https://github.com/VRCLC-DUK/ml-en-transliteration}} and an example is given in Table \ref{tab:test1}. As recommended by the task organizers, we report CER, WER, and BLEU scores separately for each test set (Table \ref{tab:final}). The distribution of these evaluation metrics over the entire test set is illustrated in Figure. \ref{fig:test1} and Figure. \ref{fig:test2}.





\section{Discussion}

In Test Set-1 with standard typing patterns, the model achieved a 7.4\% CER, demonstrating strong performance aligned with the model's training data. For Test Set-2 involving adhoc typing patterns with frequent vowel omissions, the model's performance significantly declined as indicated by the performance metrics.

While most test sentences exhibited low character-level error rates, the accompanying WER and BLEU scores appear comparatively poor. This does not indicate model inadequacy, but rather reflect the inherent limitations of WER and BLEU scores in evaluating sentence transliterations. As they penalize even minor character variations as complete word errors misrepresenting the transliteration quality \cite{james2024advocating} (Table \ref{tab:test1}).  

An error analysis exposed the model's difficulty in distinguishing phonetically similar Malayalam characters represented using same romanised form. Specifically, the model frequently misclassifies character pairs such as \includegraphics[scale=0.8]{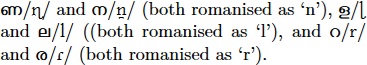}

\section{Conclusion}

Our reverse transliteration model for Malayalam demonstrates promising capabilities in converting romanized text to native script, particularly for standard typing patterns. However, the research reveals significant challenges in handling adhoc typing styles, especially those with frequent vowel omissions. Future efforts should focus on fine-tuning the model using a diverse dataset that includes a significant proportion of adhoc typing patterns to enhance its robustness and adaptability.

\section*{Limitations}

The training data primarily covers standard typing patterns and missing the nuanced variations found in irregular typing scenarios. This restricted training set significantly constrains the model's ability to generalize and accurately handle diverse input styles and patterns. Additionally, the model's design lacks a language model that could capture word dependencies and improve overall sentence-level transliteration.

\bibliography{custom}

\begin{thebibliography}{18}
\providecommand{\natexlab}[1]{#1}

\bibitem[{Baby et~al.(2016)Baby, Thomas, Nishanthi, Consortium et~al.}]{baby2016resources}
Arun Baby, Anju~Leela Thomas, NL~Nishanthi, TTS Consortium, et~al. 2016.
\newblock Resources for {I}ndian languages.
\newblock In \emph{Proceedings of Text, Speech and Dialogue}. CBBLR Workshop.

\bibitem[{Baiju et~al.(2024)Baiju, Manohar, Pillai, and Sherly}]{Baiju2024}
Bajiyo Baiju, Kavya Manohar, Leena~G Pillai, and Elizabeth Sherly. 2024.
\newblock \href {https://doi.org/10.1109/RAICS61201.2024.10690040} {{Malayalam to English Named Entity Transliteration using Attention based BiLSTM}}.
\newblock In \emph{2024 IEEE Recent Advances in Intelligent Computational Systems (RAICS)}, pages 1--6.

\bibitem[{James et~al.(2024)James, Gopinath et~al.}]{james2024advocating}
Jesin James, Deepa~P Gopinath, et~al. 2024.
\newblock Advocating character error rate for multilingual asr evaluation.
\newblock \emph{arXiv preprint arXiv:2410.07400}.

\bibitem[{Junczys-Dowmunt et~al.(2018)Junczys-Dowmunt, Grundkiewicz, Dwojak, Hoang, Heafield, Neckermann, Seide, Germann, Aji, Bogoychev, Martins, and Birch}]{junczys-dowmunt-etal-2018-marian}
Marcin Junczys-Dowmunt, Roman Grundkiewicz, Tomasz Dwojak, Hieu Hoang, Kenneth Heafield, Tom Neckermann, Frank Seide, Ulrich Germann, Alham~Fikri Aji, Nikolay Bogoychev, Andr{\'e} F.~T. Martins, and Alexandra Birch. 2018.
\newblock \href {https://doi.org/10.18653/v1/P18-4020} {{M}arian: Fast neural machine translation in {C}++}.
\newblock In \emph{Proceedings of {ACL} 2018, System Demonstrations}, pages 116--121, Melbourne, Australia. Association for Computational Linguistics.

\bibitem[{Kirov et~al.(2024)Kirov, Johny, Katanova, Gutkin, and Roark}]{kirov2024context}
Christo Kirov, Cibu Johny, Anna Katanova, Alexander Gutkin, and Brian Roark. 2024.
\newblock \href {https://doi.org/10.1162/coli_a_00510} {{Context-aware Transliteration of Romanized South Asian Languages}}.
\newblock \emph{Computational Linguistics}, pages 1--60.

\bibitem[{Kunchukuttan et~al.(2021)Kunchukuttan, Jain, and Kejriwal}]{kunchukuttan-etal-2021-large}
Anoop Kunchukuttan, Siddharth Jain, and Rahul Kejriwal. 2021.
\newblock \href {https://doi.org/10.18653/v1/2021.eacl-main.303} {A large-scale evaluation of neural machine transliteration for {I}ndic languages}.
\newblock In \emph{Proceedings of the 16th Conference of the European Chapter of the Association for Computational Linguistics: Main Volume}, pages 3469--3475, Online. Association for Computational Linguistics.

\bibitem[{Kunchukuttan et~al.(2018)Kunchukuttan, Khapra, Singh, and Bhattacharyya}]{kunchukuttan2018leveraging}
Anoop Kunchukuttan, Mitesh Khapra, Gurneet Singh, and Pushpak Bhattacharyya. 2018.
\newblock \href {https://doi.org/10.1162/tacl_a_00022} {Leveraging orthographic similarity for multilingual neural transliteration}.
\newblock \emph{Transactions of the Association for Computational Linguistics}, 6:303--316.

\bibitem[{Kunchukuttan et~al.(2015)Kunchukuttan, Puduppully, and Bhattacharyya}]{kunchukuttan-etal-2015-brahmi}
Anoop Kunchukuttan, Ratish Puduppully, and Pushpak Bhattacharyya. 2015.
\newblock \href {https://doi.org/10.3115/v1/N15-3017} {Brahmi-net: A transliteration and script conversion system for languages of the {I}ndian subcontinent}.
\newblock In \emph{Proceedings of the 2015 Conference of the North {A}merican Chapter of the Association for Computational Linguistics: Demonstrations}, pages 81--85, Denver, Colorado. Association for Computational Linguistics.

\bibitem[{Madhani et~al.(2023)Madhani, Parthan, Bedekar, Nc, Khapra, Kunchukuttan, Kumar, and Khapra}]{madhani-etal-2023-aksharantar}
Yash Madhani, Sushane Parthan, Priyanka Bedekar, Gokul Nc, Ruchi Khapra, Anoop Kunchukuttan, Pratyush Kumar, and Mitesh Khapra. 2023.
\newblock \href {https://doi.org/10.18653/v1/2023.findings-emnlp.4} {Aksharantar: Open {I}ndic-language transliteration datasets and models for the next billion users}.
\newblock In \emph{Findings of the Association for Computational Linguistics: EMNLP 2023}, pages 40--57, Singapore. Association for Computational Linguistics.

\bibitem[{Manghat et~al.(2020)Manghat, Manghat, and Schultz}]{manghat2020malayalam}
Sreeja Manghat, Sreeram Manghat, and Tanja Schultz. 2020.
\newblock \href {https://doi.org/10.21437/Interspeech.2020-1936} {{Malayalam-English Code-Switched: Grapheme to Phoneme System}}.
\newblock In \emph{Proc. Interspeech 2020}, pages 4133--4137.

\bibitem[{Manohar et~al.(2022)Manohar, Jayan, and Rajan}]{kavya2022}
Kavya Manohar, A.~R. Jayan, and Rajeev Rajan. 2022.
\newblock \href {https://doi.org/10.1109/ACCESS.2022.3204403} {Mlphon: A multifunctional grapheme-phoneme conversion tool using finite state transducers}.
\newblock \emph{IEEE Access}, 10:97555--97575.

\bibitem[{Parlikar et~al.(2016)Parlikar, Sitaram, Wilkinson, and Black}]{parlikar2016festvox}
Alok Parlikar, Sunayana Sitaram, Andrew Wilkinson, and Alan~W Black. 2016.
\newblock The {F}estvox {I}ndic frontend for grapheme to phoneme conversion.
\newblock In \emph{WILDRE: Workshop on {I}ndian Language Data-Resources and Evaluation}.

\bibitem[{Priyamvada et~al.(2022)Priyamvada, Govind, Menon, Premjith, and Soman}]{Priyamvada_2021}
R.~Priyamvada, D.~Govind, Vijay~Krishna Menon, B.~Premjith, and K.~P. Soman. 2022.
\newblock Grapheme to phoneme conversion for malayalam speech using encoder-decoder architecture.
\newblock In \emph{Intelligent Data Engineering and Analytics}, pages 41--49, Singapore. Springer Nature Singapore.

\bibitem[{Rajan(2018)}]{aksharamukha}
Vinodh Rajan. 2018.
\newblock \href {https://www.aksharamukha.com} {Aksharamukha script converter web application.}

\bibitem[{Roark et~al.(2020)Roark, Wolf-Sonkin, Kirov, Mielke, Johny, Demir{\c{s}}ahin, and Hall}]{roark-etal-2020-processing}
Brian Roark, Lawrence Wolf-Sonkin, Christo Kirov, Sabrina~J. Mielke, Cibu Johny, I{\c{s}}in Demir{\c{s}}ahin, and Keith Hall. 2020.
\newblock \href {https://www.aclweb.org/anthology/2020.lrec-1.294} {Processing {South} {Asian} languages written in the {Latin} script: the {Dakshina} dataset}.
\newblock In \emph{Proceedings of The 12th Language Resources and Evaluation Conference (LREC)}, pages 2413--2423.

\bibitem[{Thottingal(2018)}]{libindic}
Santhosh Thottingal. 2018.
\newblock \href {https://libindic.org/Transliteration} {Libindic soundex and transliteration module}.

\bibitem[{Thottingal(2023)}]{namepair}
Santhosh Thottingal. 2023.
\newblock \href {https://huggingface.co/datasets/santhosh/english-malayalam-names} {{Malayalam-English Name Pair Dataset}}.

\bibitem[{Transliteration(2001)}]{iso}
ISO~15919:2001 Transliteration. 2001.
\newblock \href {https://www.iso.org/standard/28333.html} {{Transliteration of Devanagari and related Indic scripts into Latin characters}}.

\end{thebibliography}

\appendix



\end{document}